\PassOptionsToPackage{unicode}{hyperref}
\PassOptionsToPackage{hyphens}{url}
\PassOptionsToPackage{dvipsnames,svgnames,x11names}{xcolor}
\documentclass[
  11pt]{article}
\usepackage{xcolor}
\usepackage[margin=1in]{geometry}
\usepackage{amsmath,amssymb}
\usepackage{iftex}
\ifPDFTeX
  \usepackage[T1]{fontenc}
  \usepackage[utf8]{inputenc}
  \usepackage{textcomp} % provide euro and other symbols
\else % if luatex or xetex
  \usepackage{unicode-math} % this also loads fontspec
  \defaultfontfeatures{Scale=MatchLowercase}
  \defaultfontfeatures[\rmfamily]{Ligatures=TeX,Scale=1}
\fi
\usepackage{lmodern}
\ifPDFTeX\else
\fi
\IfFileExists{upquote.sty}{\usepackage{upquote}}{}
\IfFileExists{microtype.sty}{% use microtype if available
  \usepackage[]{microtype}
  \UseMicrotypeSet[protrusion]{basicmath} % disable protrusion for tt fonts
}{}
\makeatletter
\@ifundefined{KOMAClassName}{% if non-KOMA class
  \IfFileExists{parskip.sty}{%
    \usepackage{parskip}
  }{% else
    \setlength{\parindent}{0pt}
    \setlength{\parskip}{6pt plus 2pt minus 1pt}}
}{% if KOMA class
  \KOMAoptions{parskip=half}}
\makeatother
\usepackage{color}
\usepackage{fancyvrb}

\DefineVerbatimEnvironment{Highlighting}{Verbatim}{commandchars=\\\{\}}
\newenvironment{Shaded}{}{}

\newcommand{\AttributeTok}[1]{\textcolor[rgb]{0.49,0.56,0.16}{#1}}

\newcommand{\CharTok}[1]{\textcolor[rgb]{0.25,0.44,0.63}{#1}}
\newcommand{\CommentTok}[1]{\textcolor[rgb]{0.38,0.63,0.69}{\textit{#1}}}

\newcommand{\ControlFlowTok}[1]{\textcolor[rgb]{0.00,0.44,0.13}{\textbf{#1}}}
\newcommand{\DataTypeTok}[1]{\textcolor[rgb]{0.56,0.13,0.00}{#1}}
\newcommand{\DecValTok}[1]{\textcolor[rgb]{0.25,0.63,0.44}{#1}}

\newcommand{\FunctionTok}[1]{\textcolor[rgb]{0.02,0.16,0.49}{#1}}

\newcommand{\KeywordTok}[1]{\textcolor[rgb]{0.00,0.44,0.13}{\textbf{#1}}}
\newcommand{\NormalTok}[1]{#1}
\newcommand{\OperatorTok}[1]{\textcolor[rgb]{0.40,0.40,0.40}{#1}}

\newcommand{\StringTok}[1]{\textcolor[rgb]{0.25,0.44,0.63}{#1}}

\usepackage{longtable,booktabs,array}
\usepackage{caption}
\usepackage{calc} % for calculating minipage widths
\usepackage{etoolbox}
\makeatletter
\patchcmd\longtable{\par}{\if@noskipsec\mbox{}\fi\par}{}{}
\makeatother
\IfFileExists{footnotehyper.sty}{\usepackage{footnotehyper}}{\usepackage{footnote}}
\makesavenoteenv{longtable}
\usepackage{graphicx}
\makeatletter
\newsavebox\pandoc@box
\newcommand*\pandocbounded[1]{% scales image to fit in text height/width
  \sbox\pandoc@box{#1}%
  \Gscale@div\@tempa{\textheight}{\dimexpr\ht\pandoc@box+\dp\pandoc@box\relax}%
  \Gscale@div\@tempb{\linewidth}{\wd\pandoc@box}%
  \ifdim\@tempb\p@<\@tempa\p@\let\@tempa\@tempb\fi% select the smaller of both
  \ifdim\@tempa\p@<\p@\scalebox{\@tempa}{\usebox\pandoc@box}%
  \else\usebox{\pandoc@box}%
  \fi%
}
\def\fps@figure{htbp}
\makeatother
\providecommand{\tightlist}{%
  \setlength{\itemsep}{0pt}\setlength{\parskip}{0pt}}
\usepackage{lmodern}
\usepackage{microtype}
\usepackage{amsmath,amssymb}
\usepackage{bookmark}
\IfFileExists{xurl.sty}{\usepackage{xurl}}{} % add URL line breaks if available
\makeatletter
\@ifundefined{xmpquote}{}{}
\makeatother
\hypersetup{
  pdftitle={From SQL Generation to Tool Selection: A Domain-Oriented Pattern for MCP Servers},
  pdfauthor={Bartolomeo Bogliolo},
  colorlinks=true,
  linkcolor={blue},
  filecolor={Maroon},
  citecolor={blue},
  urlcolor={blue},
  pdfcreator={LaTeX via pandoc}}

\title{From SQL Generation to Tool Selection: A Domain-Oriented Pattern
for MCP Servers}
\author{Bartolomeo Bogliolo}
\date{August 2026}

\begin{document}
\maketitle
\begin{abstract}
Agents built on Large Language Models (LLMs) increasingly reach
enterprise data through the Model Context Protocol (MCP), and many MCP
database servers maximize flexibility by exposing a single generic SQL
execution tool. This paper proposes the
\textbackslash\textbackslash textbf\{Domain-Oriented Tooling Pattern\}:
instead of generating SQL at query time, the model selects from a small
set of domain-aligned tools whose parameterized queries encapsulate
schema navigation, joins and business rules on the server side. We
formalize the pattern around three architectural invariants and
introduce \textbackslash\textbackslash textbf\{Model Demotion\}, the
observation that replacing SQL synthesis with intent classification
lowers the model tier required to serve routine requests. As a reference
implementation we present \textbackslash\textbackslash textbf\{MCP
Blueprint\}, an open-source framework that defines domain tools
declaratively as YAML metadata plus parameterized SQL files. A public
reproducibility benchmark compares three server designs --- raw SQL
execution, a thin generic pack, and a verticalized domain pack --- on
four local models (3B--8B) over seventeen enterprise-style tasks: the
verticalized pack scores 0.939 pooled mean accuracy against 0.666 for
raw SQL and 0.605 for the generic pack, with every model improving and
the smallest gaining most (0.583 → 0.929). All harness code, prompts,
gold answers, packs and per-cell results are publicly available.
\end{abstract}

\begin{center}\small
Code: \url{https://github.com/meob/mcp-blueprint} \quad
Benchmark artifacts: \url{https://github.com/meob/mcp-blueprint-benchmark} \quad
License: CC~BY~4.0
\end{center}

\section{Introduction}\label{introduction}

Large Language Models are rapidly becoming the primary interface between
users and enterprise information systems. The Model Context Protocol
(MCP) {[}1{]} accelerates this trend by standardizing how AI agents
discover and invoke external tools, and database connectivity is among
its most common applications: an MCP server fronting a relational
database lets agents answer business questions directly from operational
data.

When building such a server, developers face a design decision that
shapes everything downstream: what should the server actually expose?
The fastest path --- and a recurring pattern in published connectors ---
is a single generic tool such as \texttt{execute\_sql(query)},
optionally accompanied by schema metadata in the system prompt. This
choice maximizes short-term flexibility: the agent can explore, join and
filter anything the database credentials allow, with zero upfront domain
modeling.

It also transfers responsibilities to the model that software
engineering traditionally assigns to the application layer: schema
discovery, relationship inference, dialect handling, business-rule
interpretation, query optimization and result validation. These concerns
consume reasoning capacity, introduce non-determinism where determinism
is cheapest to guarantee, and widen the deployment's security surface.
In practice they push teams toward large, expensive frontier models
simply to compensate for an interface that exposes implementation
details instead of business concepts.

This paper argues that the abstraction layer --- not the language model
--- is the decisive design variable, and proposes replacing SQL
generation with \textbf{tool selection}: the server exposes a small set
of domain-aligned operations (``verticalized'' tools), each backed by
pre-authored parameterized SQL that encapsulates joins and business
rules.

The contributions of this paper are:

\begin{enumerate}
\def\labelenumi{\arabic{enumi}.}
\tightlist
\item
  We formalize the \textbf{Domain-Oriented Tooling Pattern} as three
  architectural invariants and relate it to established abstraction
  layers such as object-relational mapping, REST resources and semantic
  layers (Section 3).
\item
  We introduce \textbf{Model Demotion} as an engineering heuristic:
  reducing interface complexity converts open-ended SQL synthesis into
  intent classification and slot filling, lowering the model tier
  required for routine requests (Section 5).
\item
  We present \textbf{MCP Blueprint} {[}20{]}, an open-source framework
  implementing the pattern through declarative YAML/SQL pack definitions
  decoupled from protocol infrastructure.
\item
  We release a \textbf{public reproducibility benchmark} {[}21{]}
  comparing raw SQL access against two MCP tool-pack designs across four
  local models and seventeen enterprise-style tasks (Section 6). The
  verticalized pack dominates every measured dimension --- pooled
  accuracy 0.939 versus 0.666 (raw SQL) and 0.605 (generic thin-tool
  pack); the smallest 3B model matches every larger configuration; cost
  per correct answer drops by factors of roughly 2--12\(\times\). A
  superficially similar generic pack underperforms even raw SQL,
  indicating that tool \emph{design}, not tool \emph{existence}, creates
  value.
\end{enumerate}

Section 2 reviews background and motivation. Section 3 defines the
pattern, Section 4 summarizes the reference implementation, and Section
5 develops Model Demotion. Section 6 reports the benchmark design and
results, Section 7 discusses implications, Section 8 surveys related
work, and Section 9 concludes.

\begin{center}\rule{0.5\linewidth}{0.5pt}\end{center}

\section{Background and Motivation}\label{background-and-motivation}

\subsection{The Model Context
Protocol}\label{the-model-context-protocol}

The Model Context Protocol {[}1{]} standardizes how LLM applications
discover and invoke external capabilities. An MCP server advertises
typed tools --- name, description, JSON-Schema parameter contracts ---
and an MCP client surfaces these schemas to the model, which decides
when and how to call them. Transports include stdio for local processes
and Streamable HTTP for remote deployments.

Because any capability can be packaged as a tool, MCP servers now front
databases, SaaS APIs, filesystems and internal services. For relational
data specifically, the fastest way to ship a connector is to wrap a
database connection in one generic query tool and hand the schema DDL to
the model.

\subsection{Generic Query Interfaces: A Common Design
Choice}\label{generic-query-interfaces-a-common-design-choice}

A representative generic tool looks like this:

\begin{Shaded}
\begin{Highlighting}[]
\FunctionTok{\{}
  \DataTypeTok{"name"}\FunctionTok{:} \StringTok{"execute\_sql"}\FunctionTok{,}
  \DataTypeTok{"description"}\FunctionTok{:} \StringTok{"Executes an arbitrary SQL query against the target database."}\FunctionTok{,}
  \DataTypeTok{"parameters"}\FunctionTok{:} \FunctionTok{\{} \DataTypeTok{"query"}\FunctionTok{:} \StringTok{"STRING"} \FunctionTok{\}}
\FunctionTok{\}}
\end{Highlighting}
\end{Shaded}

The design is genuinely attractive at first: it requires no domain
modeling up front, covers the long tail of ad-hoc questions during
prototyping, and delegates every difficult decision to a model that is
often remarkably capable of improvising SQL.

It also transfers responsibilities that software engineering
traditionally assigns to the application layer. Consider a question such
as \emph{``Which invoices are still unpaid for customer ACME?''}
Answering it through raw SQL access requires the model to:

\begin{itemize}
\tightlist
\item
  discover which tables hold customers, invoices and payments;
\item
  infer how those entities relate and which joins apply;
\item
  find out how ``unpaid'' is represented in this particular schema;
\item
  decide which business filters must accompany the obvious ones;
\item
  produce dialect-correct, reasonably efficient SQL;
\item
  validate the shape of the returned rows before answering.
\end{itemize}

None of these steps is the user's question. They are implementation
concerns that recur --- with fresh probabilistic variation --- on every
single request.

\subsection{Operational Limitations of Generic
Interfaces}\label{operational-limitations-of-generic-interfaces}

Four consequences follow from placing these burdens on the model.

\textbf{1. Schema and context overhead.} To formulate correct SQL the
model must first ingest schema metadata. For enterprise schemas with
hundreds of tables and thousands of columns this consumes large portions
of the context window, recurs across agent iterations, inflates cost and
latency, and dilutes attention on the user's actual task.

\textbf{2. Probabilistic execution risk.} Generated queries are
synthesized anew each run and may scan unindexed tables, form Cartesian
products through missing join conditions, or fetch far more data than
needed (\texttt{SELECT\ *}). In concurrent production environments a
single pathological query can starve connection pools or induce lock
contention.

\textbf{3. Business-rule re-derivation.} Relational schemas normalize
entities and leave business semantics implicit. Whether an account
counts as ``active'', or a rental as ``overdue'', typically depends on
multi-column state evaluations and temporal comparisons encoded in
conventions rather than constraints. A model writing raw SQL must
re-derive these rules on every invocation, producing interpretations
that vary across runs, models and prompt phrasings.

\textbf{4. Enlarged security surface.} A generic execution tool grants
broad read capability by construction. Even with read-only credentials
it remains exposed to indirect prompt injection {[}17{]}, bulk data
harvesting, and resource-exhaustion patterns such as deep recursive
common table expressions. Because nothing in the interface constrains
what a query may express, every mitigation must be re-imposed at the
query boundary.

These limitations are not a verdict on text-to-SQL research, which
continues to advance rapidly {[}9{]}--{[}12{]}. They reflect deploying
an exploration-grade interface into a production serving path.

\begin{center}\rule{0.5\linewidth}{0.5pt}\end{center}

\section{The Domain-Oriented Tooling
Pattern}\label{the-domain-oriented-tooling-pattern}

\subsection{Core Philosophy}\label{core-philosophy}

The pattern is captured by a single directive:

\begin{quote}
\textbf{Do not expose the database. Expose the domain.}
\end{quote}

Under this paradigm the MCP server acts as a domain gateway. The LLM
remains an orchestrator that requests semantic information or triggers
domain actions; every data-access decision stays encapsulated inside the
server.

\begin{verbatim}
+-----------+            +---------------------+            +--------------+
| LLM Agent | --- MCP -> | Domain-Oriented     | --- SQL -> | Relational   |
| (intent)  |            | Server              |            | Database     |
+-----------+            +---------------------+            +--------------+
                                   |
                             Encapsulates:
                             - pre-authored parameterized SQL
                             - explicit tool contracts
                             - business rules
\end{verbatim}

{\def\LTcaptype{none} % do not increment counter
\begin{longtable}[]{@{}
  >{\raggedright\arraybackslash}p{(\linewidth - 4\tabcolsep) * \real{0.3333}}
  >{\raggedright\arraybackslash}p{(\linewidth - 4\tabcolsep) * \real{0.3333}}
  >{\raggedright\arraybackslash}p{(\linewidth - 4\tabcolsep) * \real{0.3333}}@{}}
\toprule\noalign{}
\begin{minipage}[b]{\linewidth}\raggedright
Dimension
\end{minipage} & \begin{minipage}[b]{\linewidth}\raggedright
Generic SQL interface
\end{minipage} & \begin{minipage}[b]{\linewidth}\raggedright
Domain-oriented interface
\end{minipage} \\
\midrule\noalign{}
\endhead
\bottomrule\noalign{}
\endlastfoot
Tool granularity & \texttt{execute\_sql()} &
\texttt{customer\_account\_summary()}, \texttt{recommend\_films()} \\
Schema knowledge & Discovered from context per request & Implicit in
reviewed tool contracts \\
Joins & Generated per request & Pre-authored and optimized \\
Business rules & Re-derived by the model & Embedded in server-side
SQL \\
Model tier & Frontier models typical & Small models frequently
sufficient \\
Output shape & Varies per run & Stable, documented columns \\
\end{longtable}
}

The transition mirrors familiar precedents. Object-relational mappers
abstracted SQL behind programming-language objects; the move from raw
RPC to RESTful resources {[}3{]} replaced unconstrained remote
invocation with bounded, self-describing operations; domain-driven
design {[}2{]} supplies the vocabulary: tools correspond to operations
of a bounded context, not to storage primitives. The Domain-Oriented
Tooling Pattern applies the same discipline to a new class of client ---
a probabilistic one that benefits even more from narrow, well-documented
contracts.

\begin{center}\rule{0.5\linewidth}{0.5pt}\end{center}

\section{MCP Blueprint: A Reference
Implementation}\label{mcp-blueprint-a-reference-implementation}

MCP Blueprint {[}20{]} operationalizes the pattern without bespoke
boilerplate for each domain. It replaces imperative server code with
declarative configuration files (``packs'') and enforces a strict
separation of concerns:

\begin{itemize}
\tightlist
\item
  \textbf{Engine layer} --- protocol serialization, stdio and Streamable
  HTTP transports, connection pooling across database engines, parameter
  validation, response caching and error handling; domain-agnostic and
  shared by all packs.
\item
  \textbf{Domain pack layer} --- YAML tool definitions, external
  parameterized SQL files and pack metadata; self-contained artifacts
  that can be versioned, reviewed, tested and ported across database
  engines.
\end{itemize}

\begin{verbatim}
mcp-blueprint/
├── blueprint/            # Core Python engine (transports, pooling, validation)
├── config/               # Server configuration (database URI, engine selection)
└── packs/
    └── sakila/           # Sakila Domain Pack
        ├── pack.yaml     # Engine compatibility & pack declaration
        ├── tools/        # Declarative YAML tool definitions
        └── sql/          # External parameterized SQL files
\end{verbatim}

Adding a domain operation is a configuration change rather than a code
change: an author writes one YAML definition and one reviewed SQL file
and commits both. Appendix A shows a complete example from the
verticalized pack used in Section 6, including how an availability rule
and an optional parameter are handled directly in SQL.

\section{Model Demotion}\label{model-demotion}

\subsection{Lowering the Cognitive
Threshold}\label{lowering-the-cognitive-threshold}

A central consequence of the pattern is what we call \textbf{Model
Demotion}: reducing interface complexity lowers the model tier required
to serve routine requests.

Through a generic SQL interface, answering a business question requires
open-ended synthesis: navigate the schema, choose join paths, translate
intent into dialect-correct SQL, execute it, inspect errors and recover.
Through a domain-oriented interface, the same request reduces to intent
classification and slot filling --- recognizing which operation applies
and extracting a few typed parameters.

\begin{verbatim}
+-------------------------------------------------------------------+
|                      GENERIC QUERY INTERFACE                      |
|   User intent -> frontier LLM -> dynamic SQL generation -> DBMS   |
+-------------------------------------------------------------------+

+-------------------------------------------------------------------+
|                     DOMAIN-ORIENTED PATTERN                       |
|   User intent -> SLM (intent/slots) -> pre-authored SQL  -> DBMS  |
+-------------------------------------------------------------------+
\end{verbatim}

The distinction matters economically and operationally:

\begin{itemize}
\tightlist
\item
  \textbf{Cost and latency} --- classifying intent over a handful of
  tools is an easier decoding problem than multi-statement code
  generation, so it can be served by small local models or inexpensive
  API tiers.
\item
  \textbf{Failure modes} --- synthesis failures are open-ended (wrong
  joins, invalid syntax, hallucinated columns); selection failures are
  bounded (wrong tool, missing parameter) and therefore easier to
  detect, log and repair.
\item
  \textbf{Determinism} --- with fixed tools, the semantic content of a
  request lives on the server rather than in model weights, so behavior
  converges across models at temperature 0.
\end{itemize}

We present Model Demotion as an engineering heuristic rather than a law:
it applies to bounded, recurring retrieval workflows, not to open-ended
analytical exploration.

\subsection{Partitioning Data Access by
Risk}\label{partitioning-data-access-by-risk}

Enterprise data requests are highly skewed: a bounded set of core
operations covers most routine traffic. Following a \textbf{95/5
heuristic}, we expect the large majority of enterprise retrieval
scenarios to be expressible as a curated collection of semantic
operations, with a long tail of exceptional requests.

The pattern partitions workload accordingly:

\begin{enumerate}
\def\labelenumi{\arabic{enumi}.}
\tightlist
\item
  \textbf{The routine majority.} Recurring operational requests are
  served deterministically by pack tools, routed by lightweight models,
  without exposing raw tables.
\item
  \textbf{The exceptional remainder.} Requests outside current coverage
  route to a human-in-the-loop workflow: a domain engineer authors,
  reviews and commits a new YAML/SQL definition, permanently absorbing
  that case into the deterministic set.
\end{enumerate}

This creates a continuous improvement cycle --- human expertise is
encoded once into a pack and then served repeatedly by cost-effective
models. Generic SQL access, by contrast, re-pays the cost of schema
understanding on every single request.

Section 6 tests the central claim of Model Demotion empirically: whether
small local models, given domain-oriented tools, can match or exceed
larger configurations working through raw SQL access.

\begin{center}\rule{0.5\linewidth}{0.5pt}\end{center}

\section{A Public Reproducibility
Benchmark}\label{a-public-reproducibility-benchmark}

To quantify the effect of interface design we built a standalone
benchmark harness and released it publicly {[}21{]}. The harness runs
identical task prompts against pluggable MCP server configurations,
scores every run against gold answers computed live from the database,
and records one JSON file per cell (model \(\times\) approach \(\times\)
task \(\times\) repetition) containing token counts, latency, agent
steps, tool-call traces and per-check results. All components needed to
reproduce the study --- harness code, task definitions, scoring rules,
gold-answer logic, the frozen packs behind approaches B and C, the DDL
used by approach A, and a run manifest pinning framework commit and pack
versions --- are committed to the repository together with the complete
per-cell record set of the reported run.

\subsection{Design and Setup}\label{design-and-setup}

\textbf{Approaches.} Three MCP server configurations expose the same
Sakila sample database (PostgreSQL port):

{\def\LTcaptype{none} % do not increment counter
\begin{longtable}[]{@{}
  >{\raggedright\arraybackslash}p{(\linewidth - 4\tabcolsep) * \real{0.3333}}
  >{\raggedright\arraybackslash}p{(\linewidth - 4\tabcolsep) * \real{0.3333}}
  >{\raggedright\arraybackslash}p{(\linewidth - 4\tabcolsep) * \real{0.3333}}@{}}
\toprule\noalign{}
\begin{minipage}[b]{\linewidth}\raggedright
\end{minipage} & \begin{minipage}[b]{\linewidth}\raggedright
Approach
\end{minipage} & \begin{minipage}[b]{\linewidth}\raggedright
Surface
\end{minipage} \\
\midrule\noalign{}
\endhead
\bottomrule\noalign{}
\endlastfoot
\textbf{A} & Raw SQL & A single \texttt{execute\_sql} tool; the DDL of
the six task-relevant tables (\texttt{customer}, \texttt{film},
\texttt{category}, \texttt{film\_category}, \texttt{inventory},
\texttt{rental}) is embedded in the system prompt. The schema fits
entirely in context, so A is not handicapped by prompt size; schema
understanding, SQL synthesis and multi-step orchestration are left to
the model. \\
\textbf{B} & Verticalized pack & MCP Blueprint loading
\texttt{packs/sakila} v0.5.0: five domain tools
(\texttt{customer\_account\_summary}, \texttt{rental\_history},
\texttt{recommend\_films}, \texttt{film\_stock},
\texttt{search\_customer}) that accept human-readable identifiers
(names, titles) and encapsulate all joins and business rules in
pre-authored SQL. \\
\textbf{C} & Generic thin-tool pack & MCP Blueprint loading a
deliberately shallow pack (\texttt{search\_customer},
\texttt{search\_films}, \texttt{get\_customer\_rentals},
\texttt{get\_film}): table-oriented tools with minimal descriptions ---
a first-pass surface a developer might ship before verticalizing. C
isolates the contribution of tool \emph{design} from tool
\emph{existence}. \\
\end{longtable}
}

\textbf{Models.} Four instruction-tuned models served locally through
Ollama:

{\def\LTcaptype{none} % do not increment counter
\begin{longtable}[]{@{}lll@{}}
\toprule\noalign{}
Model & Params & Tier \\
\midrule\noalign{}
\endhead
\bottomrule\noalign{}
\endlastfoot
llama3.2:3b & 3B & SLM \\
qwen2.5:3b & 3B & SLM \\
qwen2.5:7b & 7B & Medium \\
llama3.1:8b & 8B & Medium \\
\end{longtable}
}

Two smaller models were excluded during bring-up because their Ollama
builds do not support tool calling (HTTP 400): gemma2:2b and phi3:mini.

\textbf{Protocol.} Temperature 0; seed 42; context window 8192 tokens;
at most 10 agent steps; three repetitions per cell. The design is 4
models \(\times\) 3 approaches \(\times\) 17 tasks \(\times\) 3
repetitions = 612 planned cells, of which 609 completed (99.5\%); three
cells (qwen2.5:3b / \texttt{service\_case} / approach A) were lost to a
stdio pipe hang during MCP server startup. The run was recorded on
2026-08-20 against MCP Blueprint commit \texttt{b386d58}.

\textbf{Scoring.} Each task defines a rule-based check set evaluated
against gold answers computed live from the database; no LLM judge is
involved. Free-form titles are compared by fuzzy matching
(SequenceMatcher ratio \(\geq\) 0.72). Workflow checks (tool-call
sequences, argument validation) apply only to approaches B and C;
approach A is scored solely on its final answer text. A cell's score is
passed checks over total checks, in {[}0, 1{]}.

\subsection{Task Suite}\label{task-suite}

Seventeen customer-facing tasks stress different capabilities:

{\def\LTcaptype{none} % do not increment counter
\begin{longtable}[]{@{}
  >{\raggedright\arraybackslash}p{(\linewidth - 2\tabcolsep) * \real{0.5000}}
  >{\raggedright\arraybackslash}p{(\linewidth - 2\tabcolsep) * \real{0.5000}}@{}}
\toprule\noalign{}
\begin{minipage}[b]{\linewidth}\raggedright
Category
\end{minipage} & \begin{minipage}[b]{\linewidth}\raggedright
Tasks
\end{minipage} \\
\midrule\noalign{}
\endhead
\bottomrule\noalign{}
\endlastfoot
Customer lookup & \texttt{find\_customer}, \texttt{not\_found} \\
Rental state \& standing & \texttt{rental\_history},
\texttt{return\_verify}, \texttt{service\_case},
\texttt{overdue\_report}, \texttt{good\_standing\_recommend} \\
Recommendation & \texttt{recommend\_category},
\texttt{recommend\_rating}, \texttt{g\_available},
\texttt{avoid\_on\_loan}, \texttt{upsell\_seen}, \texttt{not\_rented} \\
Catalog details & \texttt{film\_details},
\texttt{store\_availability} \\
Multi-step workflow & \texttt{customer\_workflow} \\
Edge cases & \texttt{rental\_empty} \\
\end{longtable}
}

Negative-filtering tasks (\texttt{not\_found}, \texttt{not\_rented},
\texttt{avoid\_on\_loan}, \texttt{upsell\_seen}) and multi-step
workflows are deliberately included because they stress composition and
business-rule application rather than single-query lookup. Full prompts
appear in Appendix B.

\subsection{Accuracy Results}\label{accuracy-results}

Pooled across models, the verticalized pack dominates every accuracy
metric, while the generic pack underperforms even raw SQL:

{\def\LTcaptype{none} % do not increment counter
\begin{longtable}[]{@{}
  >{\raggedright\arraybackslash}p{(\linewidth - 6\tabcolsep) * \real{0.2500}}
  >{\raggedright\arraybackslash}p{(\linewidth - 6\tabcolsep) * \real{0.2500}}
  >{\raggedright\arraybackslash}p{(\linewidth - 6\tabcolsep) * \real{0.2500}}
  >{\raggedright\arraybackslash}p{(\linewidth - 6\tabcolsep) * \real{0.2500}}@{}}
\toprule\noalign{}
\begin{minipage}[b]{\linewidth}\raggedright
Metric
\end{minipage} & \begin{minipage}[b]{\linewidth}\raggedright
A (Raw SQL)
\end{minipage} & \begin{minipage}[b]{\linewidth}\raggedright
B (Verticalized)
\end{minipage} & \begin{minipage}[b]{\linewidth}\raggedright
C (Generic)
\end{minipage} \\
\midrule\noalign{}
\endhead
\bottomrule\noalign{}
\endlastfoot
Mean score & 0.666 & \textbf{0.939} & 0.605 \\
Fully-correct cells & 67/201 (33\%) & \textbf{174/204 (85\%)} & 63/204
(31\%) \\
Zero-score cells & 7 (3\%) & \textbf{3 (1\%)} & 19 (9\%) \\
\end{longtable}
}

Approach B holds the accuracy ceiling for every model:

{\def\LTcaptype{none} % do not increment counter
\begin{longtable}[]{@{}llll@{}}
\toprule\noalign{}
Model & A & B & C \\
\midrule\noalign{}
\endhead
\bottomrule\noalign{}
\endlastfoot
llama3.2:3b & 0.583 (6/51) & \textbf{0.929 (42/51)} & 0.419 (0/51) \\
qwen2.5:3b & 0.684 (17/48) & \textbf{0.902 (42/51)} & 0.602 (14/51) \\
qwen2.5:7b & 0.647 (20/51) & \textbf{0.958 (45/51)} & 0.631 (19/51) \\
llama3.1:8b & 0.750 (24/51) & \textbf{0.966 (45/51)} & 0.769 (30/51) \\
\end{longtable}
}

\begin{figure}
\centering
\pandocbounded{\includegraphics[width=0.92\textwidth,keepaspectratio,alt={Figure 1: Pooled mean accuracy by model and approach. The verticalized pack (B) leads at every model size, with the largest gain on the smallest model (llama3.2:3b), consistent with the Model Demotion hypothesis.}]{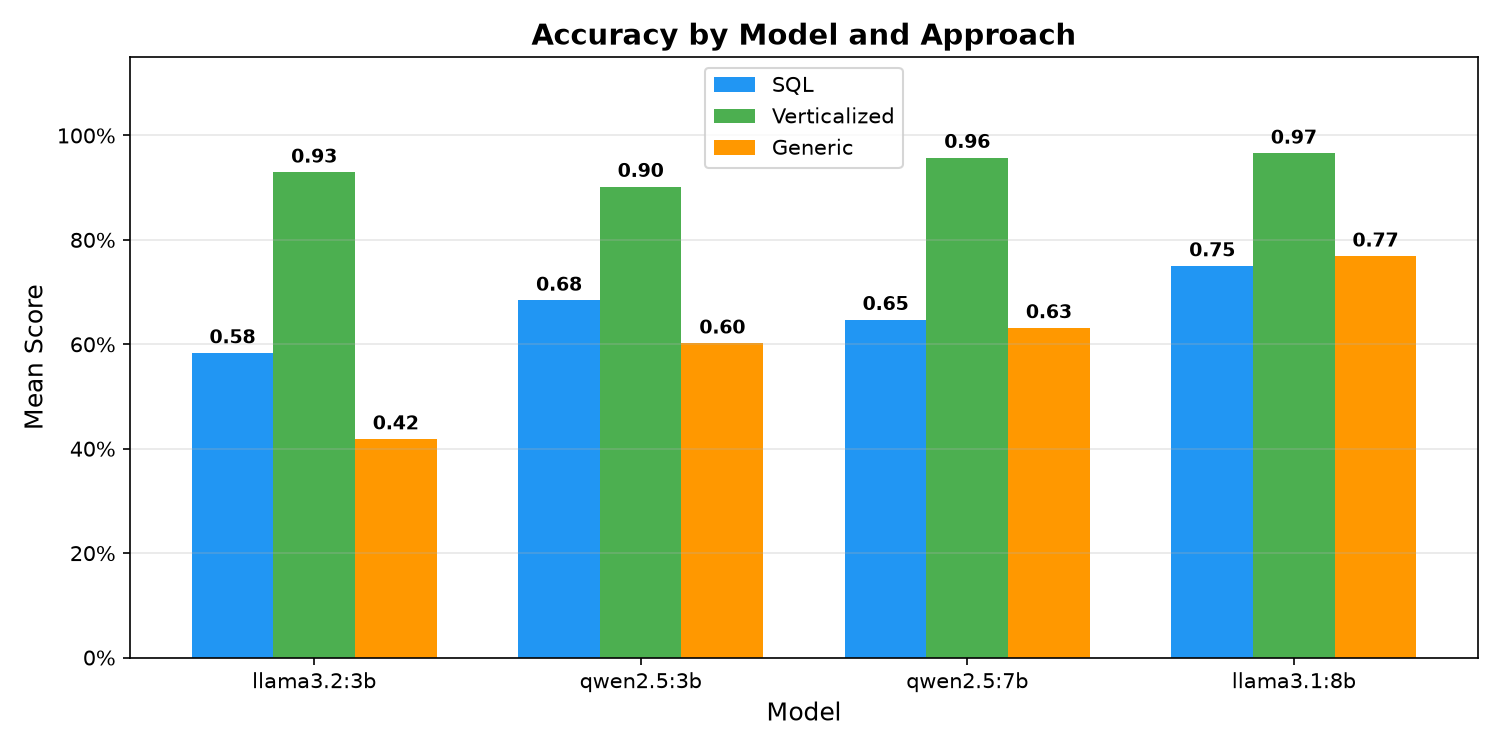}}
\caption{Pooled mean accuracy by model and approach. The
verticalized pack (B) leads at every model size, with the largest gain
on the smallest model (llama3.2:3b), consistent with the Model Demotion
hypothesis.}
\end{figure}

Three observations stand out. First, B never drops below 0.90 on any
model: the benefit does not depend on model scale. Second, the smallest
model gains the most --- llama3.2:3b improves from 0.583 under raw SQL
(with only 6/51 fully-correct cells) to 0.929 with domain tools (42/51),
which is the empirical signature of Model Demotion. Third, the generic
pack C pools below even raw SQL (0.605 vs 0.666), trailing A on three of
four models; exposing tools without designing them can be worse than not
exposing tools at all. Section 6.5 analyzes why.

\subsection{Token and Latency
Efficiency}\label{token-and-latency-efficiency}

Mean per-cell token cost is comparable across designs; the decisive
efficiency metric is cost per \emph{correct} answer, where the designs
separate sharply:

{\def\LTcaptype{none} % do not increment counter
\begin{longtable}[]{@{}
  >{\raggedright\arraybackslash}p{(\linewidth - 8\tabcolsep) * \real{0.2000}}
  >{\raggedright\arraybackslash}p{(\linewidth - 8\tabcolsep) * \real{0.2000}}
  >{\raggedright\arraybackslash}p{(\linewidth - 8\tabcolsep) * \real{0.2000}}
  >{\raggedright\arraybackslash}p{(\linewidth - 8\tabcolsep) * \real{0.2000}}
  >{\raggedright\arraybackslash}p{(\linewidth - 8\tabcolsep) * \real{0.2000}}@{}}
\toprule\noalign{}
\begin{minipage}[b]{\linewidth}\raggedright
Approach
\end{minipage} & \begin{minipage}[b]{\linewidth}\raggedright
Mean tokens/cell
\end{minipage} & \begin{minipage}[b]{\linewidth}\raggedright
vs B
\end{minipage} & \begin{minipage}[b]{\linewidth}\raggedright
Tokens per correct answer
\end{minipage} & \begin{minipage}[b]{\linewidth}\raggedright
Seconds per correct answer
\end{minipage} \\
\midrule\noalign{}
\endhead
\bottomrule\noalign{}
\endlastfoot
\textbf{B (Verticalized)} & 3,056 & --- & \textbf{3,582} &
\textbf{5.2} \\
C (Generic) & 2,894 & --5.3\% & 9,372 & 20.7 \\
A (Raw SQL) & 3,953 & +29.4\% & 11,858 & 51.9 \\
\end{longtable}
}

Mean latency per cell follows the same ordering: 4.4 s (B), 6.4 s (C)
and 17.3 s (A), with mean agent steps of 2.1, 2.4 and 2.4 and mean tool
calls of 1.1, 1.7 and 1.6 respectively. B is simultaneously the most
accurate and the fastest design in the study.

Cost per correct answer improves for every model when moving from raw
SQL to the verticalized pack:

{\def\LTcaptype{none} % do not increment counter
\begin{longtable}[]{@{}
  >{\raggedright\arraybackslash}p{(\linewidth - 10\tabcolsep) * \real{0.1667}}
  >{\raggedright\arraybackslash}p{(\linewidth - 10\tabcolsep) * \real{0.1667}}
  >{\raggedright\arraybackslash}p{(\linewidth - 10\tabcolsep) * \real{0.1667}}
  >{\raggedright\arraybackslash}p{(\linewidth - 10\tabcolsep) * \real{0.1667}}
  >{\raggedright\arraybackslash}p{(\linewidth - 10\tabcolsep) * \real{0.1667}}
  >{\raggedright\arraybackslash}p{(\linewidth - 10\tabcolsep) * \real{0.1667}}@{}}
\toprule\noalign{}
\begin{minipage}[b]{\linewidth}\raggedright
Model
\end{minipage} & \begin{minipage}[b]{\linewidth}\raggedright
A: tokens/correct
\end{minipage} & \begin{minipage}[b]{\linewidth}\raggedright
B: tokens/correct
\end{minipage} & \begin{minipage}[b]{\linewidth}\raggedright
A: s/correct
\end{minipage} & \begin{minipage}[b]{\linewidth}\raggedright
B: s/correct
\end{minipage} & \begin{minipage}[b]{\linewidth}\raggedright
Reduction
\end{minipage} \\
\midrule\noalign{}
\endhead
\bottomrule\noalign{}
\endlastfoot
llama3.2:3b & 31,476 & \textbf{2,723} & 49.5 s & \textbf{3.4 s} &
\$\approx\(11.6\)\times\$ \\
qwen2.5:3b & 17,467 & \textbf{4,766} & 125.1 s & \textbf{5.0 s} &
\$\approx\(3.7\)\times\$ \\
qwen2.5:7b & 8,464 & \textbf{4,331} & 33.2 s & \textbf{5.6 s} &
\$\approx\(2.0\)\times\$ \\
llama3.1:8b & 5,808 & \textbf{2,531} & 16.3 s & \textbf{6.7 s} &
\$\approx\(2.3\)\times\$ \\
\end{longtable}
}

The smallest model is the clearest beneficiary of Model Demotion. Under
raw SQL, llama3.2:3b consumed 3,703 tokens and 5.8 seconds per cell
while solving few cells completely; under the verticalized pack it
consumed 2,242 tokens and 2.8 seconds per cell while solving most of
them --- reducing tokens per correct answer from 31,476 to 2,723.

Equivalently: the cheapest fully-correct cell in the study is
llama3.2:3b with domain tools at roughly 2.7k tokens and 3.4 s, while
qwen2.5:7b with raw SQL spends about three times as many tokens and ten
times as long per correct answer. Equivalent quality at a fraction of
the inference budget is the economic form of the pattern's central
claim.

\subsection{Tool Design versus Tool
Existence}\label{tool-design-versus-tool-existence}

The contrast between approaches B and C is the benchmark's sharpest
finding. Both are MCP Blueprint packs exposing the same database through
parameterized tools --- yet B reaches 0.939 while C stalls at 0.605,
below even raw SQL. The difference lies entirely in how the tools are
designed:

{\def\LTcaptype{none} % do not increment counter
\begin{longtable}[]{@{}
  >{\raggedright\arraybackslash}p{(\linewidth - 4\tabcolsep) * \real{0.3333}}
  >{\raggedright\arraybackslash}p{(\linewidth - 4\tabcolsep) * \real{0.3333}}
  >{\raggedright\arraybackslash}p{(\linewidth - 4\tabcolsep) * \real{0.3333}}@{}}
\toprule\noalign{}
\begin{minipage}[b]{\linewidth}\raggedright
Dimension
\end{minipage} & \begin{minipage}[b]{\linewidth}\raggedright
B (Verticalized)
\end{minipage} & \begin{minipage}[b]{\linewidth}\raggedright
C (Generic)
\end{minipage} \\
\midrule\noalign{}
\endhead
\bottomrule\noalign{}
\endlastfoot
Tool semantics & Domain operations (\texttt{customer\_account\_summary},
\texttt{recommend\_films}) & Table-oriented (\texttt{search\_films},
\texttt{get\_customer\_rentals}) \\
Parameter contracts & Human-readable names; descriptions tuned for
matching & Bare IDs or minimal labels \\
Business logic & Encapsulated server-side (standing flag, overdue
detection) & Re-derived by the model \\
Descriptions & Rich guidance steering correct workflows & Minimal,
forcing multi-step composition \\
\end{longtable}
}

A thin tool surface leaves all reasoning with the model while removing
its freedom to compensate through arbitrary SQL: it constrains the
action space without raising answer quality. The practical implication
is that the tool surface itself --- clear parameters, explicit
contracts, logic pushed server-side --- sets the ceiling for answer
quality; neither larger models nor the mere presence of a tool layer
substitute for it.

\subsection{Per-Task Analysis}\label{per-task-analysis}

Ordering tasks by the gap between B and A shows where verticalization
contributes most:

{\def\LTcaptype{none} % do not increment counter
\begin{longtable}[]{@{}lllll@{}}
\toprule\noalign{}
Task & A & B & C & \(\Delta\)(B--A) \\
\midrule\noalign{}
\endhead
\bottomrule\noalign{}
\endlastfoot
\texttt{recommend\_category} & 0.500 & \textbf{1.000} & 0.833 &
+50.0pp \\
\texttt{avoid\_on\_loan} & 0.500 & \textbf{1.000} & 0.667 & +50.0pp \\
\texttt{upsell\_seen} & 0.500 & \textbf{1.000} & 0.750 & +50.0pp \\
\texttt{film\_details} & 0.562 & \textbf{1.000} & 0.375 & +43.8pp \\
\texttt{customer\_workflow} & 0.584 & \textbf{1.000} & 0.688 &
+41.6pp \\
\texttt{rental\_history} & 0.361 & \textbf{0.694} & 0.417 & +33.4pp \\
\texttt{store\_availability} & 0.667 & \textbf{1.000} & 0.500 &
+33.3pp \\
\texttt{good\_standing\_recommend} & 0.583 & \textbf{0.896} & 0.417 &
+31.3pp \\
\texttt{g\_available} & 0.722 & \textbf{1.000} & 0.833 & +27.8pp \\
\texttt{service\_case} & 0.694 & \textbf{0.950} & 0.350 & +25.6pp \\
\texttt{not\_rented} & 0.500 & \textbf{0.750} & 0.417 & +25.0pp \\
\texttt{return\_verify} & 0.792 & \textbf{1.000} & 0.417 & +20.8pp \\
\texttt{find\_customer} & 0.834 & \textbf{1.000} & 0.917 & +16.6pp \\
\texttt{overdue\_report} & 0.861 & \textbf{1.000} & 0.334 & +13.9pp \\
\texttt{recommend\_rating} & \textbf{1.000} & \textbf{1.000} & 0.875 &
+0.0pp \\
\texttt{not\_found} & \textbf{1.000} & \textbf{1.000} & 0.833 &
+0.0pp \\
\texttt{rental\_empty} & \textbf{0.667} & \textbf{0.667} & 0.667 &
+0.0pp \\
\end{longtable}
}

\begin{figure}
\centering
\pandocbounded{\includegraphics[width=0.92\textwidth,keepaspectratio,alt={Figure 2: Per-task mean score heatmap. Rows are tasks sorted by \textbackslash Delta(B--A); columns are model \textbackslash times approach combinations. The verticalized pack achieves near-uniform high scores; raw SQL and the generic pack degrade especially on multi-step workflow and negative-filtering tasks.}]{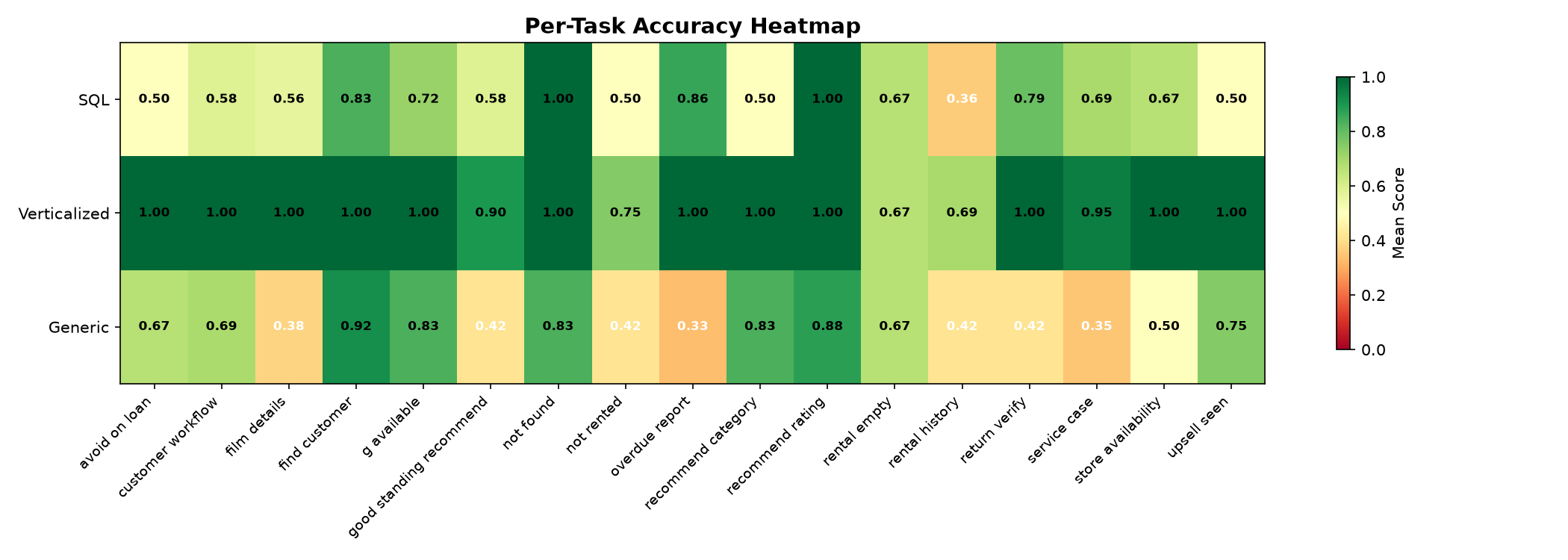}}
\caption{Per-task mean score heatmap. Rows are tasks sorted by
\(\Delta\)(B--A); columns are model \(\times\) approach combinations.
The verticalized pack achieves near-uniform high scores; raw SQL and the
generic pack degrade especially on multi-step workflow and
negative-filtering tasks.}
\end{figure}

Tasks combining recommendation with negative filtering or multi-tool
workflows (\texttt{recommend\_category}, \texttt{avoid\_on\_loan},
\texttt{upsell\_seen}, \texttt{customer\_workflow}) show gaps between
+41.6pp and +50.0pp. Tasks answerable with one simple query
(\texttt{not\_found}, \texttt{recommend\_rating}) converge at 1.000
across all approaches. The empty-result edge case \texttt{rental\_empty}
scores an identical 0.667 everywhere, suggesting that gracefully
reporting ``nothing found'' is largely model-side behavior independent
of interface design.

Under approach B, 15 of 17 tasks are fully solved (mean score 1.0) by at
least three of the four models --- twelve by all four --- indicating
that verticalized tools turn task success into a property of the
interface rather than of any particular model's SQL skill.

\subsection{Qualitative Observations}\label{qualitative-observations}

\textbf{Token economy.} Approach A embeds the DDL of six tables into
every prompt; B and C carry only concise tool descriptions. Measured
per-cell means were 3,953 tokens (A), 3,056 (B) and 2,894 (C). The
generic pack is marginally cheaper per cell than the verticalized one
--- but it answers far fewer cells correctly, so the advantage
evaporates, and inverts, once responses are weighted by correctness
(Section 6.4).

\textbf{Encapsulation of business rules.} Deciding whether an unreturned
rental is overdue requires joining \texttt{rental} to \texttt{film} and
comparing \texttt{rental\_date\ +\ rental\_duration} against the current
time --- a rule that models writing raw SQL frequently miss by checking
only \texttt{return\_date\ IS\ NULL}. Under approach B the logic lives
in \texttt{customer\_account\_summary.sql}:

\begin{Shaded}
\begin{Highlighting}[]
\ControlFlowTok{CASE}
    \ControlFlowTok{WHEN}\NormalTok{ (}
        \KeywordTok{SELECT} \FunctionTok{count}\NormalTok{(}\OperatorTok{*}\NormalTok{):}\CharTok{:int}
        \KeywordTok{FROM}\NormalTok{ rental r}
        \KeywordTok{JOIN}\NormalTok{ inventory i }\KeywordTok{ON}\NormalTok{ i.inventory\_id }\OperatorTok{=}\NormalTok{ r.inventory\_id}
        \KeywordTok{JOIN}\NormalTok{ film f }\KeywordTok{ON}\NormalTok{ f.film\_id }\OperatorTok{=}\NormalTok{ i.film\_id}
        \KeywordTok{WHERE}\NormalTok{ r.customer\_id }\OperatorTok{=}\NormalTok{ c.customer\_id}
          \KeywordTok{AND}\NormalTok{ r.return\_date }\KeywordTok{IS} \KeywordTok{NULL}
          \KeywordTok{AND}\NormalTok{ r.rental\_date }\OperatorTok{+}\NormalTok{ f.rental\_duration }\OperatorTok{*} \DataTypeTok{INTERVAL} \StringTok{\textquotesingle{}1 day\textquotesingle{}} \OperatorTok{\textless{}} \FunctionTok{CURRENT\_TIMESTAMP}
\NormalTok{    ) }\OperatorTok{\textgreater{}} \DecValTok{0} \ControlFlowTok{THEN} \StringTok{\textquotesingle{}HAS OVERDUE\textquotesingle{}}
    \ControlFlowTok{ELSE} \StringTok{\textquotesingle{}GOOD STANDING\textquotesingle{}}
\ControlFlowTok{END} \KeywordTok{AS}\NormalTok{ standing}
\end{Highlighting}
\end{Shaded}

The domain rule is guaranteed by whoever reviews the SQL file; the model
reports it instead of reconstructing it.

\textbf{Failure modes differ by interface.} Raw-SQL failures concentrate
in synthesis --- malformed joins, missing filters --- and grow as models
shrink (llama3.2:3b solved 6/51 cells perfectly under A). Generic-pack
failures concentrate in routing and workflow composition: C produced 19
zero-score cells across models, versus 7 for A and 3 for B. Under B the
argument surface reduces to a few typed parameters validated
server-side, so residual variability stems from routing mistakes rather
than query synthesis.

\textbf{Model tier economics.} The decomposition demanded by raw SQL ---
schema navigation, query synthesis, error recovery --- routinely pulls
toward larger reasoning models. Domain tools reduce the workload to
intent classification and slot filling, so equivalent service becomes
reachable by small local models: the study's cheapest correct answer
comes from a 3B model with domain tools, not from an 8B model writing
SQL.

\subsection{Threats to Validity}\label{threats-to-validity}

\begin{itemize}
\tightlist
\item
  \textbf{Task-aligned scoring.} The verticalized pack and the
  check-based scorer were developed against the same seventeen tasks.
  Part of the accuracy gap therefore reflects task-aware tool design ---
  which is intrinsic to verticalization as an engineering practice ---
  but no held-out task set was evaluated, so results should be read as
  task-aligned rather than as evidence of open-ended generalization.
\item
  \textbf{Single small schema.} Sakila exposes six task-relevant tables
  whose DDL fits entirely in context. This favors approach A if
  anything; enterprise schemas whose DDL cannot fit in context should
  widen the gap, but remain untested here.
\item
  \textbf{Local small models, single decoding setting.} Four local
  models between 3B and 8B at temperature 0 with a fixed seed. Frontier
  cloud models may narrow part of the raw-SQL gap, and robustness across
  sampling settings is unmeasured.
\item
  \textbf{Hardware-relative latency.} Latency compares approaches on
  fixed local hardware; absolute values do not transfer to other
  deployments.
\item
  \textbf{Missing and excluded runs.} Three cells (0.5\%) were lost to a
  harness fault and two models lacking tool-calling support were
  excluded before measurement; both facts are documented alongside the
  frozen results.
\end{itemize}

\begin{center}\rule{0.5\linewidth}{0.5pt}\end{center}

\section{Discussion}\label{discussion}

\textbf{For practitioners.} The benchmark suggests a concrete design
sequence for MCP database servers: inventory the recurring questions
users actually ask; expose them as named domain operations with rich
descriptions and human-readable parameters; push every join and business
rule into reviewed SQL; keep an escape hatch for uncovered requests
rather than defaulting to raw SQL access. Treat tool definitions as API
contracts --- versioned, reviewed, and covered by tests; the same rule
sets used for scoring double as regression tests for packs.

\textbf{Relation to text-to-SQL.} The pattern does not compete with
text-to-SQL research but repositions it. NL-to-SQL remains the right
interface for exploration, ad-hoc analytics and prototyping; production
serving paths benefit from bounded operations whose semantics are
guaranteed server-side. A pragmatic deployment can offer both surfaces
to different audiences under different credentials.

\textbf{When generic interfaces suffice.} Generic SQL tools remain
reasonable where data is exploratory, users are expert analysts, and
consequences are low. The pattern targets the complementary regime:
recurring operational questions, non-expert users, and autonomous agents
acting without human review of each query.

\textbf{Ecosystem implications.} Because packs are declarative
artifacts, they invite infrastructure that does not exist for
prompt-side fixes: code review workflows, portability across database
engines, role-based visibility filtering, and shared registries of
tested domain packs. Security posture also changes qualitatively: with
no arbitrary-query surface, indirect injection {[}17{]} can manipulate
routing but cannot rewrite the queries themselves.

\section{Related Work}\label{related-work}

\textbf{Agentic tool use.} ReAct {[}4{]} established interleaved
reasoning and acting as the dominant agent loop; Toolformer {[}5{]}
showed models can self-supervise tool calls, and Gorilla {[}6{]} and
ToolLLM {[}7{]} scaled tool selection over large API corpora.
Function-calling leaderboards such as BFCL {[}8{]} evaluate
tool-selection accuracy directly. This line treats the tool surface as
given; our results indicate that surface design itself is a first-order
variable worth benchmarking.

\textbf{Text-to-SQL.} Spider {[}9{]} initiated large-scale cross-domain
evaluation, followed by harder benchmarks such as BIRD {[}10{]}; DIN-SQL
{[}11{]} and DAIL-SQL {[}12{]} pushed accuracy through decomposition and
prompt engineering. These systems optimize generation quality, whereas
the Domain-Oriented Tooling Pattern removes generation from the serving
path entirely; the two approaches are complementary (Section 7).

\textbf{Abstraction layers for data access.} Object-relational mappers
abstract SQL behind language objects for imperative code; REST {[}3{]}
replaced unconstrained RPC with bounded resources; semantic layers in
analytics stacks define governed metrics. Domain-driven design {[}2{]}
supplies the organizational principle. The pattern applies this lineage
to probabilistic clients, adding requirements ORMs never faced:
descriptions must steer model routing, and parameter contracts must
tolerate natural-language inputs.

\textbf{Small open models.} Llama 3 {[}13{]}, Qwen2.5 {[}14{]}, Phi-3
{[}15{]} and Gemma 2 {[}16{]} made capable 3B--8B models runnable on
workstations. Prior work typically measures these models on
generation-heavy tasks; our benchmark measures how interface design
shifts which tier suffices for reliable tool use --- the empirical core
of Model Demotion.

\textbf{MCP ecosystem and agent reliability.} Recent work addresses
multi-agent interoperability {[}18{]}, the sustainability economics of
agentic systems {[}19{]}, and prompt-injection defenses combining nested
learning with semantic caching {[}17{]}. Our contribution is orthogonal:
it hardens the boundary between agent and database itself.

\section{Conclusion and Future Work}\label{conclusion-and-future-work}

We proposed the Domain-Oriented Tooling Pattern: expose databases to LLM
agents as curated domain operations rather than generic SQL execution.
The pattern rests on three invariants --- encapsulated data access,
deterministic business rules, declarative tool definition --- and yields
Model Demotion: when synthesis gives way to tool selection, smaller
models serve requests reliably. MCP Blueprint demonstrates that the
pattern can be implemented declaratively, keeping protocol
infrastructure separate from domain knowledge.

A public reproducibility benchmark compared three server designs across
four local models and seventeen tasks. The verticalized pack reached
0.939 pooled mean score versus 0.666 for raw SQL and 0.605 for a generic
thin-tool pack; every model gained, the smallest most of all (0.583
\(\rightarrow\) 0.929); cost per correct answer fell by factors of
roughly 2--12\(\times\); and a superficially similar generic pack scored
below raw SQL, isolating tool design --- not tool existence or model
scale --- as the decisive factor. All artifacts are public and the run
is fully reproducible.

Future work proceeds along five lines:

\begin{enumerate}
\def\labelenumi{\arabic{enumi}.}
\tightlist
\item
  \textbf{Cross-domain replication} with additional schemas, organically
  collected user requests, and held-out task protocols separating pack
  authoring from evaluation.
\item
  \textbf{Frontier and hosted models} to test whether the raw-SQL gap
  narrows at scale, and temperature sweeps for robustness
  characterization.
\item
  \textbf{Automated pack authoring}: generating candidate YAML/SQL
  definitions from validated views and OpenAPI specifications, with
  human review as the acceptance gate.
\item
  \textbf{Governance features}: role-aware tool visibility, audit
  trails, and pack registries with compatibility certification.
\item
  \textbf{Beyond relational stores}: extending the declarative pack
  model to document stores and vector retrieval systems.
\end{enumerate}

\begin{center}\rule{0.5\linewidth}{0.5pt}\end{center}

\appendix

\section{Declarative Tool
Specification}\label{declarative-tool-specification}

A complete domain tool consists of one YAML file and one SQL file. The
YAML declares the tool name, its natural-language description (the
primary routing signal for the model), typed parameters, the backing SQL
file, and cache behavior:

\begin{Shaded}
\begin{Highlighting}[]
\CommentTok{\# packs/sakila/tools/film\_stock.yaml}
\FunctionTok{name}\KeywordTok{:}\AttributeTok{ film\_stock}
\FunctionTok{description}\KeywordTok{: }\CharTok{\textgreater{}{-}}
\NormalTok{  Per{-}store stock for a film found by title.  One call returns one row per}
\NormalTok{  store with the total copies, the copies currently available (not on loan),}
\NormalTok{  plus the film\textquotesingle{}s rating and length in minutes.  The title is matched}
\NormalTok{  case{-}insensitively as a substring.  Pass store\_id only to filter to a}
\NormalTok{  single store.  Use this when a customer asks how many copies of a film are}
\NormalTok{  available or whether it is in stock at a store.}
\FunctionTok{parameters}\KeywordTok{:}
\AttributeTok{  }\FunctionTok{title}\KeywordTok{:}
\AttributeTok{    }\FunctionTok{type}\KeywordTok{:}\AttributeTok{ string}
\AttributeTok{    }\FunctionTok{required}\KeywordTok{:}\AttributeTok{ }\CharTok{true}
\AttributeTok{    }\FunctionTok{description}\KeywordTok{:}\AttributeTok{ Film title or a substring of it, matched case{-}insensitively.}
\AttributeTok{  }\FunctionTok{store\_id}\KeywordTok{:}
\AttributeTok{    }\FunctionTok{type}\KeywordTok{:}\AttributeTok{ integer}
\AttributeTok{    }\FunctionTok{required}\KeywordTok{:}\AttributeTok{ }\CharTok{false}
\AttributeTok{    }\FunctionTok{default}\KeywordTok{:}\AttributeTok{ }\CharTok{null}
\AttributeTok{    }\FunctionTok{description}\KeywordTok{:}\AttributeTok{ Optional store identifier to filter the result to one store.}
\FunctionTok{sql}\KeywordTok{:}\AttributeTok{ ../sql/film\_stock.sql}
\FunctionTok{cache}\KeywordTok{:}
\AttributeTok{  }\FunctionTok{ttl}\KeywordTok{:}\AttributeTok{ }\DecValTok{30}
\end{Highlighting}
\end{Shaded}

SQL files remain isolated from application code. Values always arrive as
bound placeholders (\texttt{\%(title)s}, \texttt{\%(store\_id)s}),
preventing injection, and the business rule --- a copy is available when
no open rental references it --- is computed inside the query itself
rather than by the model. Optional parameters are handled with a
lightweight template conditional (\texttt{\{\%\ if\ store\_id\ \%\}}):
when the model omits \texttt{store\_id}, the filter simply disappears
from the rendered SQL, so one definition serves both filtered and
unfiltered calls.

\begin{Shaded}
\begin{Highlighting}[]
\CommentTok{{-}{-} packs/sakila/sql/film\_stock.sql}
\KeywordTok{SELECT}\NormalTok{ f.film\_id,}
\NormalTok{       f.title,}
\NormalTok{       f.rating:}\CharTok{:text}  \KeywordTok{AS}\NormalTok{ rating,}
\NormalTok{       f.}\FunctionTok{length}\NormalTok{,}
\NormalTok{       s.store\_id,}
       \FunctionTok{COUNT}\NormalTok{(i.inventory\_id)  }\KeywordTok{AS}\NormalTok{ total\_copies,}
       \FunctionTok{COUNT}\NormalTok{(i.inventory\_id) }\KeywordTok{FILTER}\NormalTok{ (}
           \KeywordTok{WHERE} \KeywordTok{NOT} \KeywordTok{EXISTS}\NormalTok{ (}
               \KeywordTok{SELECT} \DecValTok{1}
               \KeywordTok{FROM}\NormalTok{ rental r}
               \KeywordTok{WHERE}\NormalTok{ r.inventory\_id }\OperatorTok{=}\NormalTok{ i.inventory\_id}
                 \KeywordTok{AND}\NormalTok{ r.return\_date }\KeywordTok{IS} \KeywordTok{NULL}
\NormalTok{           )}
\NormalTok{       )  }\KeywordTok{AS}\NormalTok{ available}
\KeywordTok{FROM}\NormalTok{ film f}
\KeywordTok{JOIN}\NormalTok{ inventory i }\KeywordTok{ON}\NormalTok{ i.film\_id }\OperatorTok{=}\NormalTok{ f.film\_id}
\KeywordTok{JOIN} \KeywordTok{store}\NormalTok{ s }\KeywordTok{ON}\NormalTok{ s.store\_id }\OperatorTok{=}\NormalTok{ i.store\_id}
\KeywordTok{WHERE}\NormalTok{ f.title ILIKE }\StringTok{\textquotesingle{}\%\%\textquotesingle{}} \OperatorTok{||}\NormalTok{ \%(title)s }\OperatorTok{||} \StringTok{\textquotesingle{}\%\%\textquotesingle{}}
\NormalTok{\{\% }\ControlFlowTok{if}\NormalTok{ store\_id \%\}}
  \KeywordTok{AND}\NormalTok{ s.store\_id }\OperatorTok{=}\NormalTok{ \%(store\_id)s}
\NormalTok{\{\% endif \%\}}
\KeywordTok{GROUP} \KeywordTok{BY}\NormalTok{ f.film\_id, f.title, f.rating, f.}\FunctionTok{length}\NormalTok{, s.store\_id}
\KeywordTok{ORDER} \KeywordTok{BY}\NormalTok{ f.film\_id, s.store\_id}
\KeywordTok{LIMIT} \DecValTok{50}\NormalTok{;}
\end{Highlighting}
\end{Shaded}

Because both files are declarative artifacts, the tool can be reviewed,
tested, versioned and ported to another database engine without touching
protocol code.

\begin{center}\rule{0.5\linewidth}{0.5pt}\end{center}

\section{Benchmark Tasks}\label{benchmark-tasks}

{\def\LTcaptype{none} % do not increment counter
\begin{longtable}[]{@{}
  >{\raggedright\arraybackslash}p{(\linewidth - 4\tabcolsep) * \real{0.1429}}
  >{\raggedright\arraybackslash}p{(\linewidth - 4\tabcolsep) * \real{0.4286}}
  >{\raggedright\arraybackslash}p{(\linewidth - 4\tabcolsep) * \real{0.4286}}@{}}
\toprule\noalign{}
\begin{minipage}[b]{\linewidth}\raggedright
\#
\end{minipage} & \begin{minipage}[b]{\linewidth}\raggedright
Task id
\end{minipage} & \begin{minipage}[b]{\linewidth}\raggedright
Prompt
\end{minipage} \\
\midrule\noalign{}
\endhead
\bottomrule\noalign{}
\endlastfoot
1 & \texttt{find\_customer} & Find the customer whose last name is
Smith. Report the customer's full name and customer ID. \\
2 & \texttt{rental\_history} & Mary Smith wants to know what she has
rented. What films has she rented, and does she currently have any
rentals outstanding? \\
3 & \texttt{good\_standing\_recommend} & Check whether customer Maria
Miller has any overdue rentals. If she is in good standing, recommend
two Sci-Fi movies. \\
4 & \texttt{overdue\_report} & Check whether customer Tammy Sanders has
any overdue rentals. If she does, list the films she still has to
return. \\
5 & \texttt{recommend\_category} & Recommend three popular Family movies
for a family movie night. \\
6 & \texttt{recommend\_rating} & Recommend two movies rated G, suitable
for all ages. \\
7 & \texttt{film\_details} & Tell me about the movie `Goodfellas
Salute': its rating, its length in minutes, and how many copies are
available. \\
8 & \texttt{avoid\_on\_loan} & Customer Tammy Sanders is at the counter
right now. Recommend one Science Fiction movie that she is not currently
renting. \\
9 & \texttt{not\_found} & Find the customer whose last name is Doe. \\
10 & \texttt{customer\_workflow} & A customer named Jennifer Davis is
asking about her account. Check her rental situation and then recommend
a Documentary movie she might enjoy. \\
11 & \texttt{upsell\_seen} & Customer Kelly Torres enjoyed the Science
Fiction movies she has rented before. Recommend two other popular
Science Fiction movies that she has NOT rented before. \\
12 & \texttt{return\_verify} & Customer Mary Smith says she has returned
everything she rented. Verify from the records: does she still have any
rentals on loan? If so, list the film titles. \\
13 & \texttt{store\_availability} & A customer at Store 2 is asking for
`Goodfellas Salute'. How many copies are available, and is it in stock
at Store 2? \\
14 & \texttt{g\_available} & Recommend a G-rated movie that is currently
available in stock for a family movie night. \\
15 & \texttt{service\_case} & Customer Tammy Sanders is calling about
late fees. Check her account: which films are currently on loan, which
of those are overdue, and what is her home store so the store can reach
out to her? \\
16 & \texttt{rental\_empty} & Show me the rental history for customer ID
9999. \\
17 & \texttt{not\_rented} & Which Science Fiction movies has Mary Smith
NOT rented before? \\
\end{longtable}
}

\begin{center}\rule{0.5\linewidth}{0.5pt}\end{center}

\section*{References}

\begin{enumerate}
\def\labelenumi{\arabic{enumi}.}
\tightlist
\item
  Anthropic. (2024). \emph{Model Context Protocol Specification}.
  https://modelcontextprotocol.io
\item
  Evans, E. (2004). \emph{Domain-Driven Design: Tackling Complexity in
  the Heart of Software}. Addison-Wesley.
\item
  Fielding, R. T. (2000). \emph{Architectural Styles and the Design of
  Network-based Software Architectures} (Doctoral dissertation,
  University of California, Irvine).
\item
  Yao, S., Zhao, J., Yu, D., Du, N., Shafran, I., Narasimhan, K., \&
  Cao, Y. (2023). \emph{ReAct: Synergizing Reasoning and Acting in
  Language Models}. ICLR. arXiv:2210.03629.
\item
  Schick, T., Dwivedi-Yu, J., Dessì, R., Raileanu, R., Lomeli, M.,
  Zettlemoyer, L., Cancedda, N., \& Scialom, T. (2023).
  \emph{Toolformer: Language Models Can Teach Themselves to Use Tools}.
  NeurIPS. arXiv:2302.04761.
\item
  Patil, S. G., Zhang, T., Wang, X., \& Gonzalez, J. E. (2023).
  \emph{Gorilla: Large Language Model Connected with Massive APIs}.
  arXiv:2305.15334.
\item
  Qin, Y., Liang, S., Ye, Y., Zhu, K., Yan, L., Lu, Y., Lin, Y., Cong,
  X., Tang, X., Qian, B., et al.~(2023). \emph{ToolLLM: Facilitating
  Large Language Models to Master 16000+ Real-world APIs}.
  arXiv:2307.16789.
\item
  Yan, F., Mao, H., Liu, C. C., Tang, K., Lou, R., Li, H., Yin, W., Yu,
  P. S., \& Zhang, T. (2024). \emph{Berkeley Function Calling
  Leaderboard}. arXiv:2408.04682.
\item
  Yu, T., Zhang, R., Yang, K., Yasunaga, M., Wang, D., Li, Z., et
  al.~(2018). \emph{Spider: A Large-Scale Human-Labeled Dataset for
  Complex and Cross-Domain Semantic Parsing and Text-to-SQL Task}.
  EMNLP. arXiv:1809.08887.
\item
  Li, J., Yuan, Y., Zhang, G., Yu, B., Li, L., Li, T., et al.~(2023).
  \emph{Can LLM Already Serve as a Database Interface? A BI Benchmark on
  Complex SQLs (BIRD)}. NeurIPS Datasets and Benchmarks.
  arXiv:2305.03111.
\item
  Pourreza, M., \& Rafiei, D. (2023). \emph{DIN-SQL: Decomposed
  In-Context Learning of Text-to-SQL with Self-Correction}. NeurIPS.
  arXiv:2304.11015.
\item
  Gao, D., Wang, H., Li, Y., Sun, X., Qian, Y., Bian, B., et al.~(2024).
  \emph{Text-to-SQL Empowered by Large Language Models: A Benchmark
  Evaluation (DAIL-SQL)}. PVLDB 17(5). arXiv:2308.15363.
\item
  Meta AI. (2024). \emph{The Llama 3 Herd of Models}. arXiv:2407.21783.
\item
  Qwen Team. (2024). \emph{Qwen2.5 Technical Report}. arXiv:2412.15115.
\item
  Abdin, M., et al.~(2024). \emph{Phi-3 Technical Report: A Highly
  Capable Language Model Locally on Your Phone}. arXiv:2404.14219.
\item
  Gemma Team. (2024). \emph{Gemma 2: Improving Open Language Models at a
  Practical Size}. arXiv:2408.00118.
\item
  Gosmar, D., \& Dahl, D. A. (2026). \emph{Prompt Injection Mitigation
  with Agentic AI, Nested Learning, and AI Sustainability via Semantic
  Caching}. IFIP AIAI, Springer. doi:10.1007/978-3-032-30805-4\_21
\item
  Gosmar, D., Dahl, D. A., Coin, E., \& Attwater, D. (2024). \emph{AI
  Multi-Agent Interoperability Extension for Managing Multiparty
  Conversations}. arXiv:2411.05828.
\item
  Gosmar, D., Pallotta, A. C., \& Zenezini, G. (2025). \emph{Agentic AI
  Sustainability Assessment for Supply Chain Document Insights}.
  arXiv:2511.07097.
\item
  Bogliolo, B. (2026). \emph{MCP Blueprint: Declarative Framework for
  Model Context Protocol Servers}. https://github.com/meob/mcp-blueprint
\item
  Bogliolo, B. (2026). \emph{MCP Blueprint Benchmark: Reproducible
  Harness for MCP Server Design Evaluation}.
  https://github.com/meob/mcp-blueprint-benchmark
\end{enumerate}

\end{document}